\documentclass[runningheads]{llncs}
\newcommand{\comment}[1]{}

\newcommand{\etal}{\emph{et al. }}

\newcommand{\by}{\mathbf{y}}

\newcommand{\bA}{\mathbf{A}}

\newcommand{\bC}{\mathbf{C}}

\newcommand{\bI}{\mathbf{I}}

\newcommand{\bY}{\mathbf{Y}}

\newcommand{\mM}{\mathcal{M}}

\newcommand{\mL}{\mathcal{L}}

\usepackage{graphicx}
\usepackage{amsmath,amssymb} % define this before the line numbering.
\usepackage{color}
\usepackage[width=122mm,left=12mm,paperwidth=146mm,height=193mm,top=12mm,paperheight=217mm]{geometry}
\usepackage{hyperref}
\usepackage{xcolor}
\definecolor{orange}{cmyk}{0,0.6,1,0}
\definecolor{ForestGreen}{rgb}{0.0, 0.5, 0.0}

\begin{document}
% \linenumbers
\pagestyle{headings}
\mainmatter

% \title{Anatomically Coherent \\Facial Expression Synthesis}
\title{GANimation: Anatomically-aware Facial Animation from a Single Image}
% Replace with your title

\titlerunning{GANimation: Anatomically-aware Facial Animation from a Single Image}

% Replace with a meaningful short version of your title

\authorrunning{Pumarola \textit{et al.}}
% Replace with shorter version of the author list. If there are more authors than fits a line, please use A. Author et al.

\author{Albert Pumarola$^{1}$, 
Antonio Agudo$^{1}$,
Aleix M. Martinez$^{2}$, \\
Alberto Sanfeliu$^{1}$,
Francesc Moreno-Noguer$^{1}$}

%Please write out author names in full in the paper, i.e. full given and family names. 
%If any authors have names that can be parsed into FirstName LastName in multiple ways, please include the correct parsing, in a comment to the volume editors:
%\index{Lastnames, Firstnames}
%(Do not uncomment it, because you may introduce extra index items if you do that, we will use scripts for introducing index entries...)

\institute{$^{1}$Institut de Rob\`otica i Inform\`atica Industrial, CSIC-UPC, 08028, Barcelona, Spain\\
$^{2}$The Ohio State University, Columbus, OH 43210, USA\\
}

\maketitle

\begin{abstract}
Recent advances in Generative Adversarial Networks (GANs) have shown impressive results for  task of facial expression synthesis. The most successful architecture is StarGAN~\cite{choi2017stargan},   that conditions GANs'  generation process with images of a specific domain, namely a set of images of persons sharing the same expression. While effective, this approach can only generate a discrete number of expressions, determined by the  content of the dataset. To address this limitation, in this paper, we introduce a novel GAN conditioning scheme based on  Action Units (AU) annotations, which describes in a continuous manifold  the  anatomical facial  movements defining a human expression. Our approach allows controlling the magnitude of activation of each AU and combine several of them. Additionally, we propose a fully unsupervised strategy to train the model, that only requires  images annotated with their activated AUs, and exploit attention mechanisms that make our network robust to changing backgrounds and lighting conditions. Extensive evaluation show that our approach goes beyond competing conditional generators both in the capability to synthesize a much wider range of expressions ruled by anatomically feasible muscle movements, as in the capacity of dealing with images in the wild. 
\keywords{GANs,  Face Animation, Action-Unit Condition. }
\end{abstract}

\section{Introduction}
Being able to automatically animate the facial expression from a single image would open  the door to many new exciting applications in different areas, including the movie industry, photography technologies, fashion and e-commerce business, to name but a few. As Generative and Adversarial Networks have become more prevalent, this task has experienced significant advances, with architectures such as StarGAN~\cite{choi2017stargan},  which is able not only to synthesize novel expressions, but also to change other attributes of the face, such as  age, hair color or gender. Despite its generality, StarGAN can only change a particular aspect of a face among a discrete   number of attributes defined by the annotation granularity of the dataset. For instance, for the facial expression synthesis task, \cite{choi2017stargan}  is trained on the RaFD~\cite{langner2010presentation} dataset  which has only 8 binary labels for facial expressions, namely sad, neutral, angry, contemptuous, disgusted, surprised, fearful and happy.

Facial expressions, however, are the result of the combined and coordinated action of facial muscles that cannot be categorized in a discrete and low number of classes.  
Ekman and Friesen~\cite{Ekman1978} developed the Facial Action Coding System (FACS) for describing facial expressions in terms of the so-called Action Units (AUs), which are anatomically related to the contractions of specific facial muscles. Although the number of action units is relatively small (30 AUs were found to be anatomically related to the contraction of specific facial muscles), more than 7,000 different AU combinations have been observed~\cite{Scherer1982}. For example, the facial expression for \textit{fear} is generally produced with activations: Inner Brow Raiser (AU1), Outer Brow Raiser (AU2), Brow Lowerer (AU4), Upper Lid Raiser (AU5), Lid Tightener (AU7), Lip Stretcher (AU20) and Jaw Drop (AU26)~\cite{du2014compound}. Depending on the magnitude of each AU, the expression will transmit  the emotion of fear to a greater or lesser extent.

\begin{figure}[t!]
\hspace{-0.4cm}
\resizebox{12.7cm}{!} {
\includegraphics[width=\linewidth]{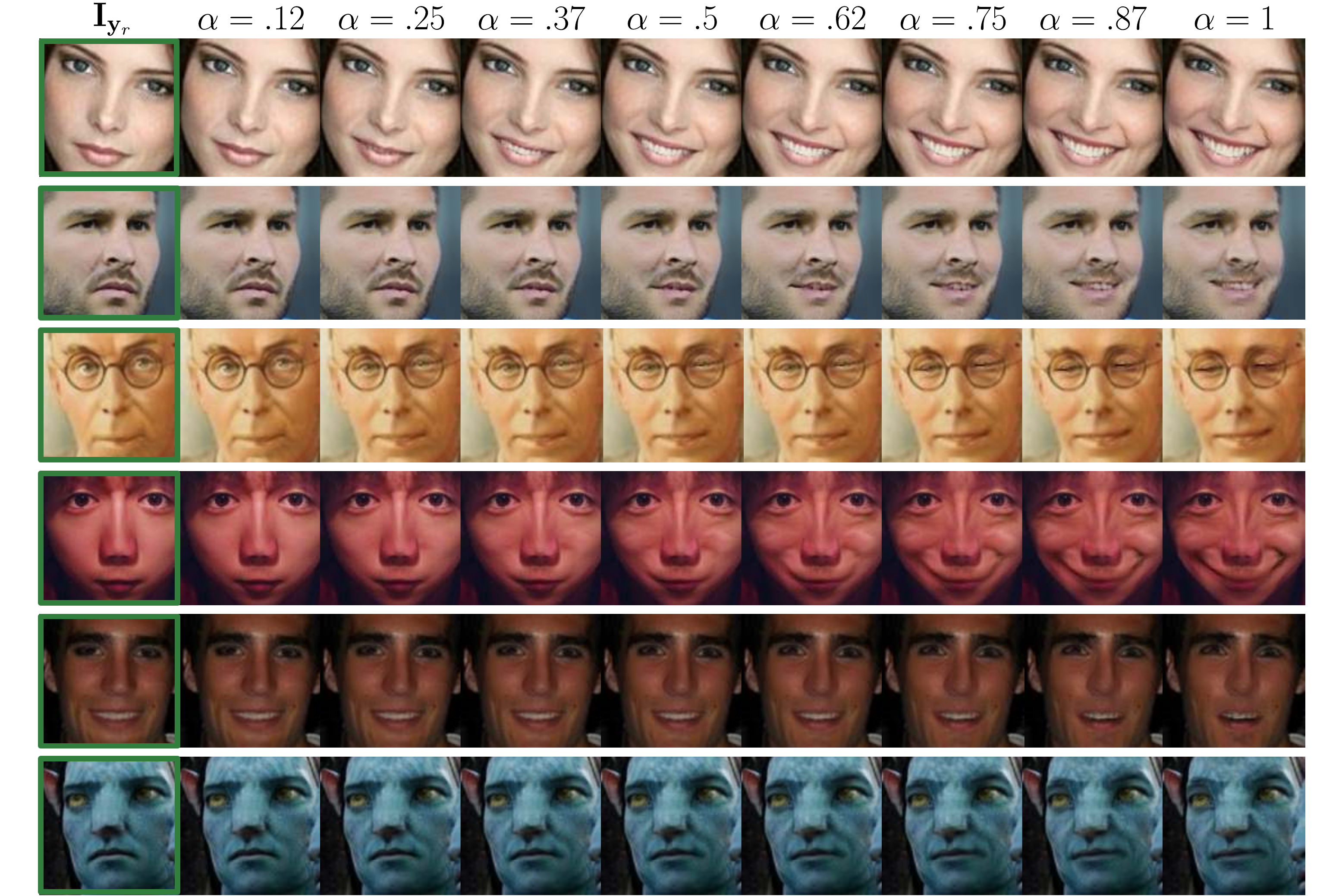}}
\vspace{-7mm}
\caption{\textbf{Facial animation from a single image}. We propose an anatomically coherent approach that is not constrained to a discrete number of expressions and can animate a given image and render novel expressions in a continuum. In these examples, we are given solely the left-most input image $\bI_{\by_r}$ (highlighted by a green square), and the parameter $\alpha$ controls the degree of activation of the target action units involved in a smiling-like expression. Additionally, our system can handle images with unnatural illumination conditions, such as the example in the bottom row.}
\vspace{-4mm}
\label{fig:intro}
\end{figure}

In this paper we aim at building a model for synthetic facial animation with the level of expressiveness of FACS, and being able to generate anatomically-aware expressions in a continuous domain, without the need of obtaining any facial landmarks~\cite{zafeiriou2017menpo}. For this purpose we leverage on the recent EmotioNet dataset~\cite{benitez2016emotionet}, which consists of one million images of facial expressions (we use 200,000 of them) of emotion in the wild annotated with discrete AUs activations~\footnote{The dataset was re-annotated with~\cite{baltruvsaitis2015cross} to obtain continuous activation annotations.}. We build a GAN architecture which, instead of being conditioned with images of a specific domain as in~\cite{choi2017stargan}, it is conditioned on a one-dimensional vector indicating the presence/absence and the magnitude of each action unit. We train this architecture in an unsupervised manner that only requires images with their activated AUs. To circumvent the need for pairs of training images of the same person under different expressions, we split the problem in two main stages. First, we consider an AU-conditioned bidirectional adversarial architecture which, given a single training photo, initially renders a new image under the desired expression. This synthesized image is then rendered-back to the original pose, hence being directly comparable to the input image. We incorporate  very recent losses to   assess the photorealism of the generated image.
Additionally, our system also goes beyond state-of-the-art in that it can handle images under changing backgrounds and illumination conditions. We achieve this by means of an attention layer that focuses the action of the network only in those regions of the image that are relevant to convey the novel expression.

As a result, we build an anatomically coherent facial expression synthesis method, able to render images in a continuous domain, and which can handle images in the wild with complex backgrounds and illumination conditions. As we will show in the results section, it compares favorably to other conditioned-GANs schemes, both in terms of the visual quality of the results, and the possibilities of generation. Figure~\ref{fig:intro} shows some example of the results we obtain, in which given one input image, we gradually change the magnitude of activation of the AUs used to produce a smile.

\section{Related Work}

\noindent{\bf Generative Adversarial Networks.} 
GANs are a powerful class of generative models based on game theory. A typical GAN optimization consists in simultaneously training a generator network to produce realistic fake samples and a discriminator network trained to distinguish between real and fake data. This idea is embedded by the so-called {\em adversarial loss}. Recent works~\cite{arjovsky2017wasserstein,gulrajani2017improved} have shown improved stability relaying on the continuous Earth Mover Distance metric, which we shall use in this paper to train our model. GANs have been shown to produce very realistic images with a high level of detail and have been successfully used for image translation~\cite{zhu2017unpaired,isola2016image,kim2017learning}, face generation~\cite{karras2017progressive,radford2015unsupervised} , super-resolution imaging~\cite{wang2015deep,ledig2016photo}, indoor scene modeling~\cite{karras2017progressive,wang2016generative} and human poses editing~\cite{pumarola2018usupervised}. 

\vspace{1mm}
\noindent{\bf Conditional GANs.} 
An active area of research is designing GAN models that incorporate conditions and constraints into the generation process. Prior studies have explored combining several conditions, such as text descriptions~\cite{reed2016generative,zhu2017your,zhang2017stackgan} and class information~\cite{odena2016conditional,mirza2014conditional}. Particularly interesting for this work are those methods exploring image based conditioning as in image super-resolution~\cite{ledig2016photo}, future frame prediction~\cite{mathieu2016deep}, image in-painting~\cite{pathak2016context}, image-to-image translation~\cite{isola2016image} and multi-target domain transfer~\cite{choi2017stargan}.

\vspace{1mm}
\noindent{\bf Unpaired Image-to-Image Translation.} 
As in our framework, several works have also tackled the problem of using unpaired training data. First attempts~\cite{liu2017unsupervised} relied on Markov random field priors for Bayesian based generation models using images from the marginal distributions in individual domains. Others explored enhancing GANS with Variational Auto-Encoder strategies~\cite{liu2017unsupervised,kingma2013auto}. Later, several works~\cite{pathak2016context,li2016precomputed} have exploited the idea of driving the system to produce mappings transforming the style without altering the original input image content. Our approach is more related to those works exploiting cycle consistency to preserve key attributes between the input and the mapped image, such as CycleGAN~\cite{zhu2017unpaired}, DiscoGAN~\cite{kim2017learning} and StarGAN~\cite{choi2017stargan}.

\vspace{1mm}
\noindent{\bf Face Image Manipulation.} Face generation and editing is a well-studied topic in computer vision and generative models. Most works have tackled the task on attribute editing~\cite{larsen2015autoencoding,perarnau2016invertible,shen2017learning} trying to modify attribute categories such as adding glasses, changing color hair, gender swapping and aging. The works that are most related to ours are those synthesizing facial expressions.  Early approaches addressed the problem using mass-and-spring models to physically approximate skin and muscle movement~\cite{FischlerTC1973}. The problem with this approach is that is difficult to generate natural looking facial expressions as there are many subtle skin movements that are difficult to render with simple spring models. Another line of research relied on 2D and 3D morphings~\cite{yu2012perception}, but produced strong artifacts around the region boundaries and was not able to model  illumination changes.

More recent works~\cite{choi2017stargan,odena2016conditional,li2016deep} train highly complex convolutional networks able to work with images in the wild. However, these approaches have been conditioned on discrete emotion categories (e.g., happy, neutral, and sad). Instead, our model resumes the idea of modeling skin and muscles, but we integrate it in modern deep learning machinery. More specifically, we learn a GAN model conditioned on a continuous embedding of muscle movements, allowing to generate a large range of  anatomically possible face expressions as well as smooth facial movement transitions in video sequences.

%--------------------------------------------------------
\section{Problem Formulation}
Let us define an input RGB image as $\bI_{\by_r} \in \mathbb{R}^{H \times W \times 3}$, captured under an arbitrary facial expression. Every gesture expression is encoded by means of a set of $N$ action units $\by_r=(y_1,\ldots,y_{N})^\top$, where each $y_n$ denotes a normalized value between 0 and 1 to module the magnitude of the $n$-th action unit. It is worth pointing out that thanks to this continuous representation, a natural interpolation can be done between different expressions, allowing to render a wide range of realistic and smooth facial expressions. 

Our aim is to learn a mapping $\mM$ to translate $\bI_{\by_r}$ into an output image $\bI_{\by_g}$ conditioned on an action-unit target $\by_g$, i.e., we seek to estimate the mapping $\mM: (\bI_{\by_r},\by_g) \rightarrow \bI_{\by_g}$. To this end, we propose to train $\mM$ in an unsupervised manner, using  $M$ training triplets $\{ \bI_{\by_r}^m,\by_r^m,\by_g^m\}_{m=1}^M$, where the target vectors $\by_g^m$ are randomly generated. Importantly, we neither   require pairs of images of the same person under different expressions, nor the expected target image $\bI_{\by_g}$.

\begin{figure}[t!]
\includegraphics[width=\linewidth]{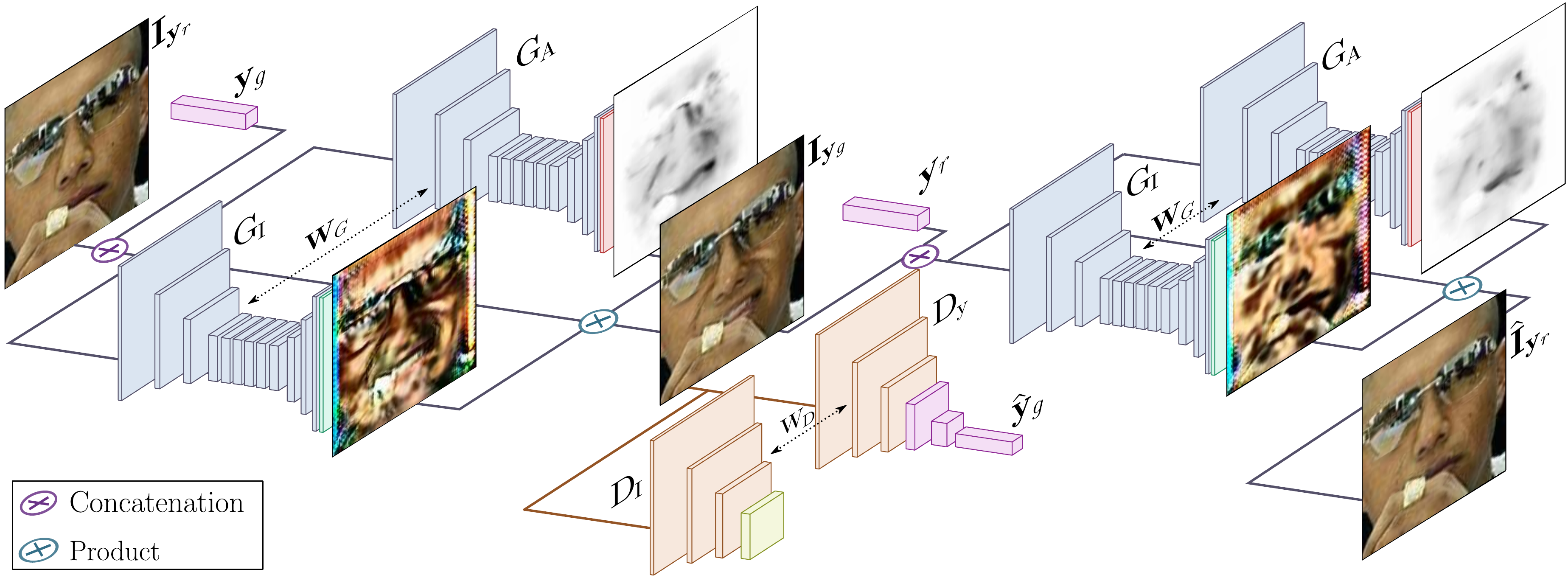}
\vspace{-7mm}
\caption{\textbf{Overview of our  approach to generate photo-realistic conditioned images.} The proposed architecture consists of two main blocks: a generator $G$ to regress   attention and color masks; and a critic $D$ to evaluate the generated image in its photorealism $D_I$ and expression conditioning fulfillment $\hat{\by}_g$. Note that our systems does not require supervision, i.e.,  no pairs of images of the same person with different expressions, nor the target image $\bI_{\by_g}$ are assumed to be known. }
\vspace{-2mm}
\label{fig:overal_model}
\end{figure}

%------------------------------------------
\section{Our Approach}

This section describes  our novel approach to generate photo-realistic conditioned images, which, as shown in Fig.~\ref{fig:overal_model},   consists of two main modules. On the one hand, a generator $G(\bI_{\by_r}|\by_g)$ is trained to realistically transform the facial expression in image $\bI_{\by_r}$ to the desired $\by_g$. Note that $G$ is applied twice, first to map the input image $\bI_{\by_r}\rightarrow \bI_{\by_g}$, and then to render it back $\bI_{\by_g}\rightarrow \hat{\bI}_{\by_r}$. On the other hand, we use a WGAN-GP~\cite{gulrajani2017improved} based critic $D(\bI_{\by_g})$ to evaluate the quality of the generated image as well as its expression.

%-----------------------------------------------------
\subsection{Network Architecture}

\vspace{1mm}
\noindent{\bf Generator.}
Let  $G$ be the generator block. Since it will be applied bidirectionally (i.e., to map either input image to desired expression and vice-versa) in the following discussion we use subscripts $o$ and $f$ to indicate {\em origin} and {\em final}.

Given the image $\bI_{\by_o}\in\mathbb{R}^{H \times W \times 3}$ and the $N$-vector $\by_f$ encoding the desired expression, we form  the input of generator as a concatenation  $(\bI_{\by_o},\by_o)\in\mathbb{R}^{H \times W \times (N+3)}$, where $\by_o$ has been represented as $N$ arrays of size $H\times W$. 

\begin{figure}[t!]
\centering
\includegraphics[width=\linewidth]{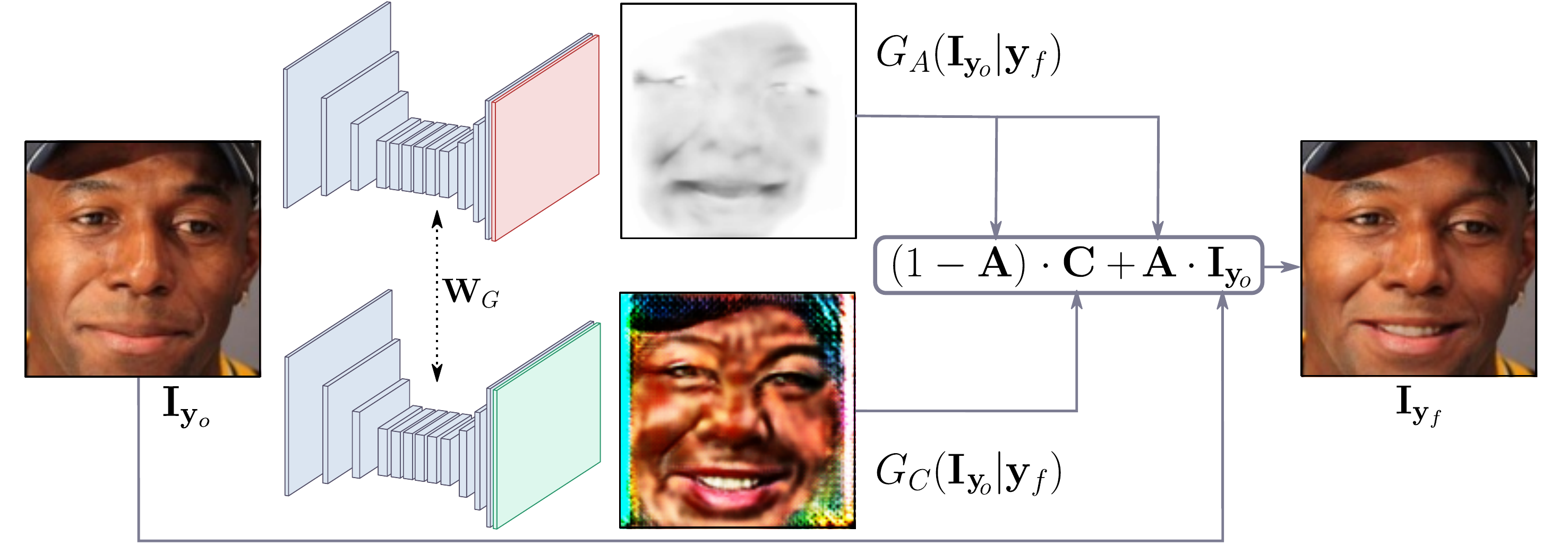}
\vspace{-7mm}
\caption{\textbf{Attention-based generator.} Given an input image and the target expression, the generator regresses and attention mask $\bA$ and an RGB color transformation $\bC$ over the entire image. The attention mask defines a per pixel intensity specifying to which extend each pixel of the original image will contribute in the final rendered image.} 
\vspace{-2mm}
\label{fig:attention_eq}
\end{figure}

One key ingredient of our system is to make $G$ focus only on those regions of the image that are responsible of synthesizing the novel expression and keep the rest elements of the image such as hair, glasses, hats or jewelery untouched. For this purpose, we have embedded an attention mechanism to the generator. Concretely, instead of regressing a full image, our generator outputs two masks, a color mask $\bC$ and attention mask $\bA$. The final image can be obtained as:

\begin{equation}
\label{eq:merge}
\bI_{\by_f} = (1 - \bA) \cdot \bC + \bA \cdot \bI_{\by_o} \,\, ,
\end{equation}
where $\bA=G_A(\bI_{\by_o}|\by_f)\in \{0,\ldots,1\}^{H \times W}$ and $\bC=G_C(\bI_{y_o}|\by_f)\in\mathbb{R}^{H \times W \times 3}$. The mask $\bA$ indicates to which extend each pixel of the $\bC$ contributes to the output image $\bI_{\by_f}$. In this way, the generator does not need to render static elements, and can focus exclusively on the pixels defining the facial movements, leading to sharper and more realistic synthetic images. This process is depicted in Fig.~\ref{fig:attention_eq}.

\vspace{1mm}
\noindent{\bf Conditional Critic.}
This is a network trained to evaluate the generated images in terms of their photo-realism and desired expression fulfillment. The structure of $D(\bI)$ resembles that of  the PatchGan~\cite{isola2016image} network mapping from the input image $\bI$ to a matrix  $\bY_{\text{I}}\in\mathbb{R}^{H/2^6 \times W/2^6}$, where $\bY_{\text{I}}[i,j]$  represents the probability of the overlapping patch $ij$ to be real. Also, to evaluate its conditioning, on top of it we add an auxiliary regression head that estimates the AUs activations $\hat{\by}=(\hat{y}_1,\ldots,\hat{y}_{N})^\top$  in the image.

\subsection{Learning the Model}

The loss function we define contains four terms, namely an \textit{image adversarial loss}~\cite{arjovsky2017wasserstein} with the modification proposed by Gulrajani \etal~\cite{gulrajani2017improved} that pushes the distribution of the generated images to the distribution of the training images; the \textit{attention loss} to drive the attention masks to be smooth and prevent them from saturating; the \textit{conditional expression loss} that conditions the expression of the generated images to be similar to the desired one; and the \textit{identity loss} that favors to preserve the person texture identity.

\vspace{1mm}
\noindent{\bf Image Adversarial Loss.}
In order to learn the parameters of the generator $G$, we use the modification of the standard GAN algorithm~\cite{goodfellow2014generative} proposed by WGAN-GP~\cite{gulrajani2017improved}. Specifically, the original GAN formulation is based on the Jensen-Shannon (JS) divergence loss function and aims to maximize the probability of correctly classifying real and rendered images while the generator tries to foul the discriminator. This loss is potentially not continuous with respect to the generator’s parameters and can locally saturate leading to vanishing gradients in the discriminator. This is addressed in WGAN~\cite{arjovsky2017wasserstein} by replacing JS with the continuous Earth Mover Distance. To maintain a Lipschitz constraint, WGAN-GP~\cite{gulrajani2017improved} proposes to add a gradient penalty for the critic network computed as the norm of the gradients with respect to the critic input.

Formally, let $\bI_{\by_o}$ be the input image with the initial condition $\by_o$, $\by_f$ the desired final condition, $\mathbb{P}_o$ the data distribution of the input image, and $\mathbb{P}_{\widetilde{I}}$ the random interpolation distribution. Then, the \textit{critic loss} $\mathcal{L}_{\text{I}}(G, D_{\text{I}}, \bI_{\by_o}, \by_f)$ we use is:
\begin{align}
&  \mathbb{E}_{\bI_{\by_o} \sim \mathbb{P}_o} [ D_{\text{I}}(G(\bI_{\by_o}|\by_f)) ] - \mathbb{E}_{\bI_{\by_o} \sim \mathbb{P}_o} \left[ D_{\text{I}}(\bI_{\by_o}) \right]+ \lambda_{\text{gp}} \mathbb{E}_{\widetilde{I} \sim \mathbb{P}_{\widetilde{I}}} \left[ ( \| \nabla_{\widetilde{I}} D_{\text{I}}(\widetilde{I}) \|_2-1)^2\right], \nonumber
\end{align}
where $\lambda_{\text{gp}}$ is a penalty coefficient. 

\vspace{1mm}
\noindent{\bf Attention Loss.}
When training the model we do not have ground-truth annotation for the attention masks $\bA$. Similarly as for the color masks $\bC$, they are learned from  the resulting gradients of the critic module and the rest of the losses. However, the attention masks can easily saturate to $1$ which makes that $\bI_{\by_o} = G(\bI_{\by_o}|\by_f)$, that is, the generator has no effect. To prevent this situation, we regularize the mask with a $l_2$-weight penalty. Also, to enforce smooth spatial color transformation when combining the pixel from the input image and the color transformation $\bC$, we perform a \textit{Total Variation Regularization} over $\bA$. The attention loss $\mathcal{L}_{\text{A}}(G, \bI_{\by_o}, \by_f)$ can therefore be defined as:
\begin{align}
%&\mathcal{L}_{\text{A}}(G, \bI_{\by_o}, \by_f)=\\
& \lambda_{\text{TV}} \mathbb{E}_{\bI_{\by_o} \sim \mathbb{P}_o} \left [\sum_{i,j}^{H,W} \left [(\bA_{i+1,j} - \bA_{i,j} )^{2} + (\bA_{i,j+1} - \bA_{i,j} )^{2} \right ] \right ] + \mathbb{E}_{\bI_{\by_o} \sim \mathbb{P}_o} \left[\| \bA \|_2\right] 
\end{align}
where $\bA = G_A(\bI_{\by_o}|\by_f)$ and $\bA_{i,j}$ is the $i,j$ entry of $\bA$. $\lambda_{\text{TV}}$ is a penalty coefficient.

\vspace{1mm}
\noindent{\bf Conditional Expression Loss.}
While reducing the \textit{image adversarial loss}, the generator must also reduce the error produced by the AUs regression head on top of $D$. In this way, $G$ not only learns to render realistic samples but also learns to satisfy the target facial expression encoded by $\by_f$. This loss is defined with two components: an AUs regression loss with fake images used to optimize G, and an AUs regression loss of real images used to learn the regression head on top of D. This loss $\mathcal{L}_{\text{y}}(G, D_{\text{y}}, \bI_{\by_o}, \by_o, \by_f)$ is computed as:
\begin{align}
%&\mathcal{L}_{\text{y}}(G, D_{\text{y}}, \bI_{\by_o}, \by_o, \by_f)= \\
& \mathbb{E}_{\bI_{\by_o} \sim \mathbb{P}_o} \left[\| D_{\text{y}}(G(\bI_{\by_o}|\by_f))] - \by_f \|^2_2\right] + \mathbb{E}_{\bI_{\by_o} \sim \mathbb{P}_o} \left[\| D_{\text{y}}(\bI_{\by_o}) - \by_o \|^2_2\right]. 
\end{align}
 
 \vspace{1mm}
\noindent{\bf Identity Loss.}
 With the previously defined losses the generator is enforced to generate photo-realistic face transformations. However, without ground-truth supervision, there is no constraint to guarantee that the face in both the input and output images correspond to the same person. Using a \textit{cycle consistency loss}~\cite{zhu2017unpaired} we force the generator to maintain the identity of each individual by penalizing the difference between the original image $\bI_{\by_o}$ and its reconstruction:
\begin{equation}
 \mathcal{L}_{\text{idt}}(G, \bI_{\by_o}, \by_o, \by_f) = \mathbb{E}_{\bI_{\by_o} \sim \mathbb{P}_o} \left[\| G(G(\bI_{\by_o}|\by_f)|\by_o) - \bI_{\by_o} \|_1\right] .
\end{equation}
To produce realistic images it is critical for the generator to model both low and high frequencies. Our \textit{PatchGan} based critic $D_{\text{I}}$ already enforces high-frequency correctness by restricting  our attention to the structure in local image patches.
To also capture low-frequencies it is sufficient to use $l_1$-norm. In preliminary experiments, we also tried replacing $l_1$-norm with a more sophisticated \textit{Perceptual}~\cite{johnson2016perceptual} loss, although we did not observe improved performance. 
 
\vspace{1mm}
\noindent{\bf Full Loss.} 
To generate the target image $\bI_{\by_g}$, we build a loss function $\mL$ by linearly combining all previous partial losses:
\begin{align}
\mL =& \mathcal{L}_{\text{I}}(G, D_{\text{I}}, \bI_{\by_r}, \by_g) + \lambda_{\text{y}} \mathcal{L}_{\text{y}}(G, D_{\text{y}}, \bI_{\by_r}, \by_r, \by_g) \label{eq:fullloss} \\ &+ \lambda_{\text{A}} \left( \mathcal{L}_{\text{A}}(G, \bI_{\by_g}, \by_r) + \mathcal{L}_{\text{A}}(G, \bI_{\by_r}, \by_g) \right) + \lambda_{\text{idt}}\mathcal{L}_{\text{idt}}(G, \bI_{\by_r}, \by_r, \by_g) ,\nonumber
\end{align}
where $\lambda_{\text{A}}$, $\lambda_{\text{y}}$ and $\lambda_{\text{idt}}$ are the hyper-parameters that  control the relative importance of every loss term. Finally, we can define the following minimax problem: 
\begin{equation}
G^\star =\arg \min_{G} \max_{D \in \mathcal{D}} \mathcal{L} \,\, ,
 \end{equation} 
where $G^\star$ draws samples from the data distribution. Additionally, we constrain our discriminator $D$ to lie in $\mathcal{D}$, that represents the set of 1-Lipschitz functions.

\section{Implementation Details}
Our generator builds upon  the variation of the network from Johnson~\etal~\cite{johnson2016perceptual} proposed by~\cite{zhu2017unpaired} as it proved to achieve impressive results for image-to-image mapping. We have slightly modified it by substituting the last convolutional layer with two parallel convolutional layers, one to regress the color mask $\bC$ and the other to define the attention mask $\bA$. We also observed that changing batch normalization in the generator by instance normalization improved training stability. For the critic we have adopted the \textit{PatchGan} architecture of~\cite{isola2016image},  but removing feature normalization. Otherwise, when computing the gradient penalty, the norm of the critic's gradient would be computed with respect to the entire batch and not with respect to each input independently. 

\begin{figure}[t!]
\centering
\includegraphics[width=\linewidth]{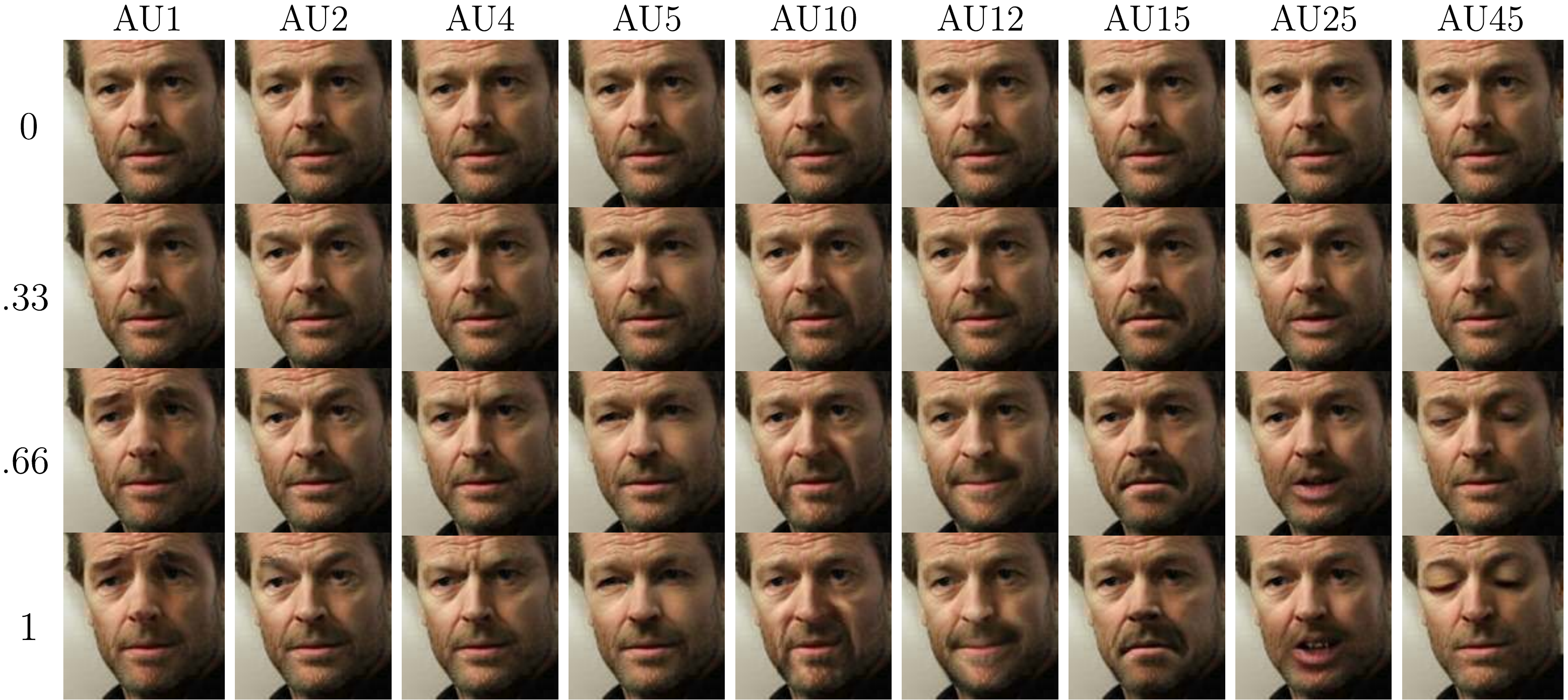}
\vspace{-7mm}
\caption{\textbf{Single AUs edition.} Specific AUs are activated at increasing levels of intensity (from 0.33 to 1). The first row corresponds to a zero intensity application of the AU which correctly produces the original image in all cases.}
\label{fig:au_editing1}
\vspace{-2mm}
\end{figure}

The model is trained on the EmotioNet dataset~\cite{benitez2016emotionet}. We use a subset of 200,000 samples (over 1 million)  to reduce  training time. We  use Adam~\cite{kingma2014adam} with learning rate of 0.0001, beta1 0.5, beta2 0.999 and batch size 25. We train for 30 epochs and linearly decay the rate to zero over the last 10 epochs. Every 5 optimization steps of the critic network we perform a single optimization step of the generator. The weight coefficients for the loss terms in Eq.~\eqref{eq:fullloss} are set to $\lambda_{\text{gp}}=10$, $\lambda_{\text{A}}=0.1$, $\lambda_{\text{TV}}=0.0001$, $\lambda_{\text{y}}=4000$, $\lambda_{\text{idt}}=10$. To improve stability we tried updating the critic using a buffer with generated images in different updates of the generator as proposed in~\cite{shrivastava2016learning} but we did not observe performance improvement. The model takes two days to train with a single GeForce\textsuperscript{\textregistered} GTX 1080 Ti GPU.

\section{Experimental Evaluation}
This section provides a thorough evaluation of our system. We first  test the main component, namely the single and multiple AUs editing. We then compare our model against current competing techniques in the task of discrete emotions editing and demonstrate our model's ability to deal with images in the wild and its capability to generate a wide range of anatomically coherent face transformations. Finally, we discuss the model's limitations and failure cases.

It is worth noting that in some of the experiments the input faces are not cropped. In this cases we first use a   detector~\footnote{We use the face detector from \url{https://github.com/ageitgey/face_recognition.}} to localize and crop the face, apply the expression transformation to that area with Eq.~\eqref{eq:merge}, and finally place the generated face back to its original position in the image. The attention mechanism guaranties a smooth transition between the morphed cropped face and the original image. As we shall see later, this three steps process results on higher resolution images compared to previous models. 
Supplementary material can be found on  \url{http://www.albertpumarola.com/research/GANimation/}.

\subsection{Single Action Units Edition}

We first evaluate our model's ability to activate AUs at different intensities while preserving the person's identity. Figure~\ref{fig:au_editing1} shows a subset of 9 AUs individually transformed with four levels of intensity (0, 0.33, 0.66, 1). For the case of 0 intensity it is desired not to change the corresponding AU. The model properly handles this situation and generates an identical copy of the input image for every case. The ability to apply an identity transformation is essential to ensure that non-desired facial movement will not be introduced. 

\begin{figure}[t!]
\centering
\includegraphics[width=\linewidth]{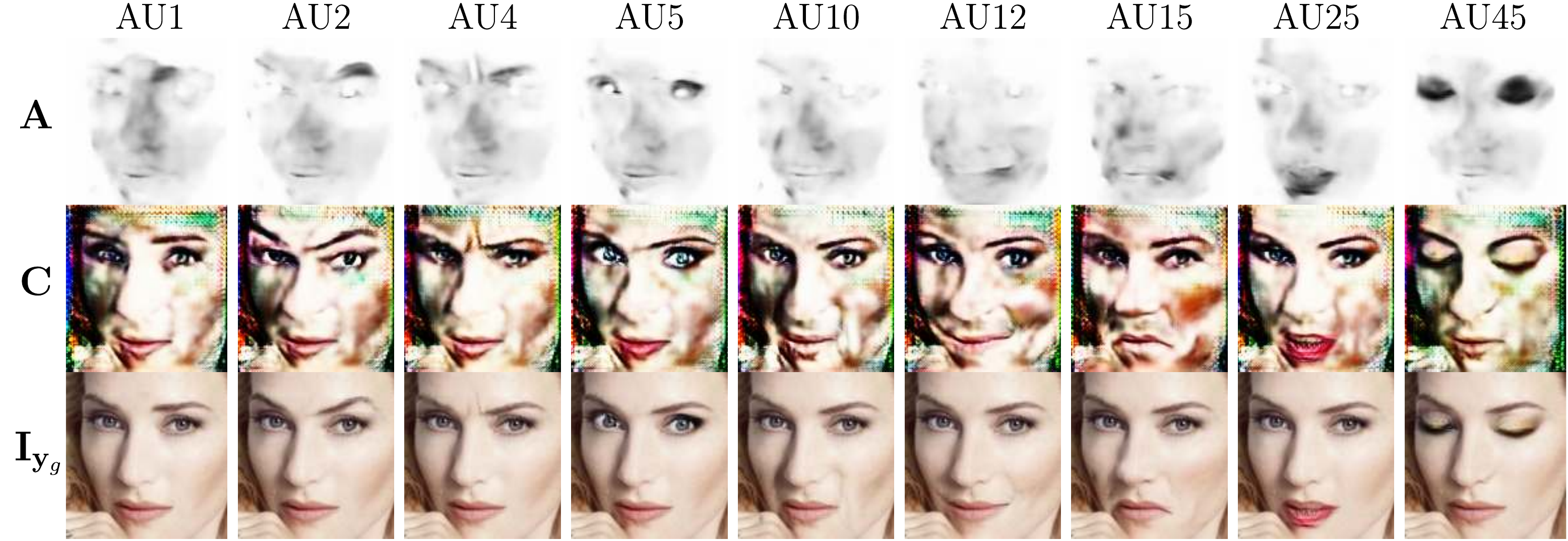}
\vspace{-7mm}
\caption{\textbf{Attention Model.} Details of the intermediate attention mask $\bA$ (first row) and the color mask $\bC$ (second row). The bottom row images are the synthesized expressions. Darker regions of the attention mask $\bA$ show those areas of the image more relevant for each specific AU. Brighter areas are retained from the original image.}
\vspace{-3mm}
\label{fig:au_editing2}
\end{figure}

For the non-zero cases, it can be observed how each AU is progressively accentuated. Note the difference between generated images at intensity 0 and 1. The model convincingly renders complex facial movements which in most cases are difficult to distinguish from real images. It is also worth mentioning that the independence of facial muscle cluster is properly learned by the generator. AUs relative to the eyes and half-upper part of the face (AUs 1, 2, 4, 5, 45) do not affect the muscles of the mouth. Equivalently, mouth related transformations (AUs 10, 12, 15, 25) do not affect eyes nor eyebrow muscles. 

Fig.~\ref{fig:au_editing2} displays, for the same experiment, the attention $\bA$ and color $\bC$ masks that produced the final result $\bI_{\by_g}$. Note how the model has learned to focus its attention (darker area) onto the corresponding AU in an unsupervised manner. In this way, it  relieves the color mask from having to accurately regress each pixel value. Only the pixels relevant to the expression change are carefully   estimated, the rest are just noise. For example, the attention is clearly obviating background pixels allowing to directly copy them from the original image. This is a   key ingredient to later being able to handle  images in the wild (see Section~\ref{sec:wild}).

\subsection{Simultaneous Edition of Multiple AUs}
We next push  the limits of our model and evaluate it in editing multiple AUs. Additionally, we also assess its ability to interpolate between two expressions. The results of this experiment are shown in Fig.~\ref{fig:intro}, the first column is the original image with expression $\by_r$, and the  right-most column is a synthetically generated image conditioned on a target expression $\by_g$. The rest of columns result from evaluating the generator conditioned with a linear interpolation of the original  and target expressions: $\alpha \by_g + (1-\alpha) \by_r$. The outcomes show a very remarkable smooth an consistent transformation across frames. We have intentionally selected challenging samples to show robustness to light conditions and even, as in the case of the avatar, to non-real world data distributions which were not previously seen by the model. These results are encouraging to further extend the model to video generation in future works.

\begin{figure}[t!]
\centering
\includegraphics[width=\linewidth]{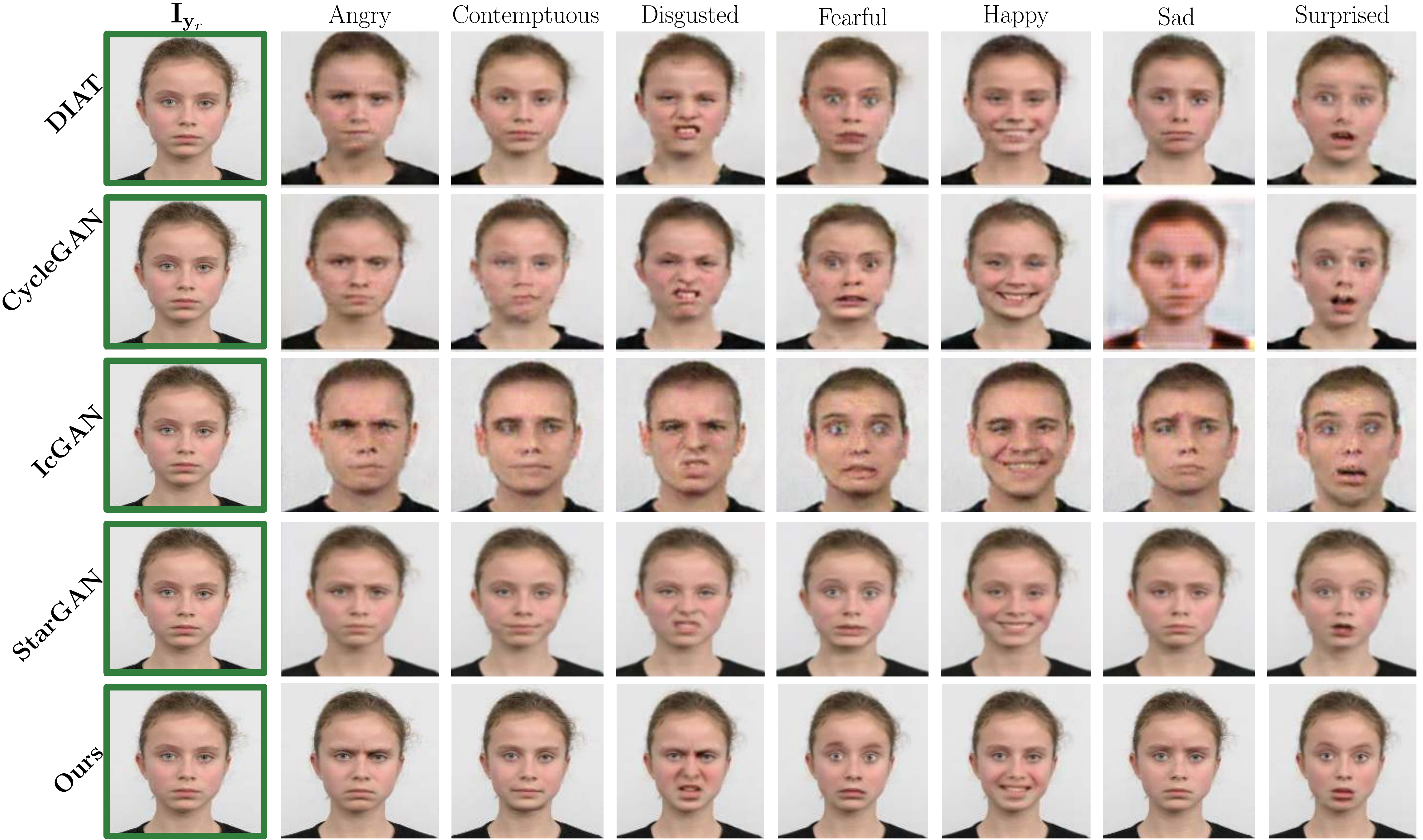}
\vspace{-7mm}
\caption{\textbf{Qualitative comparison with state-of-the-art.} Facial Expression Synthesis results for: DIAT~\cite{li2016deep}, CycleGAN~\cite{radford2015unsupervised}, IcGAN~\cite{perarnau2016invertible} and StarGAN~\cite{choi2017stargan}; and ours. In all cases, we represent the input image and seven different facial expressions. As it can be seen, our solution produces the best trade-off between visual accuracy and spatial resolution. Some of the results of StarGAN, the best current approach, show certain level of blur.  Images of previous models were taken from~\cite{choi2017stargan}.}
\vspace{-2mm}
\label{fig:state-of-the-art-comp}
\end{figure}

\subsection{Discrete Emotions Editing}
We next compare our approach, against  the baselines DIAT~\cite{li2016deep}, CycleGAN~\cite{radford2015unsupervised}, IcGAN~\cite{perarnau2016invertible} and StarGAN~\cite{choi2017stargan}. For a fair comparison, we adopt the results of these methods trained by the most recent work, StarGAN, on the task of rendering discrete emotions categories (e.g., happy, sad and fearful) in the RaFD dataset~\cite{langner2010presentation}. Since DIAT~\cite{li2016deep} and CycleGAN~\cite{radford2015unsupervised} do not allow conditioning, they were independently trained for every possible pair of source/target emotions. We next briefly discuss the main aspects of each approach:

\vspace{1mm}
\noindent{DIAT~\cite{li2016deep}.} Given an input image $x \in X$ and a reference image $y \in Y$, DIAT learns a GAN model to render the attributes of domain $Y$ in the image $x$  while conserving the person's identity. It is trained with the classic \textit{adversarial loss} and a \textit{cycle loss} $\|x - G_{Y \rightarrow X}(G_{X \rightarrow Y}(x))\|_1$ to preserve the person's identity.

\vspace{1mm}
\noindent{CycleGAN~\cite{radford2015unsupervised}.} Similar to DIAT~\cite{li2016deep}, CycleGAN also learns the mapping between two domains $X \rightarrow Y$ and $Y \rightarrow X$. To train the domain transfer, it uses a regularization term denoted \textit{cycle consistency loss} combining two cycles: $\|x - G_{Y \rightarrow X}(G_{X \rightarrow Y}(x))\|_1$ and $\|y - G_{X \rightarrow Y}(G_{Y \rightarrow X}(y))\|_1$.

\vspace{1mm}
\noindent{IcGAN~\cite{perarnau2016invertible}.} Given an input image, IcGAN uses a pretrained encoder-decoder to encode the image into a latent representation in concatenation with an expression vector $\by$ to then reconstruct the original image. It can modify the expression by replacing $\by$ with the desired expression before going through the decoder.

\vspace{1mm}
\noindent{StarGAN}~\cite{choi2017stargan}. An extension of \textit{cycle loss} for simultaneously
training between multiple datasets with different data domains. It uses a mask vector to ignore unspecified labels and optimize only on known ground-truth labels. It yields more realistic results when training simultaneously with multiple datasets.

\begin{figure}[t!]
\centering
\includegraphics[width=\linewidth]{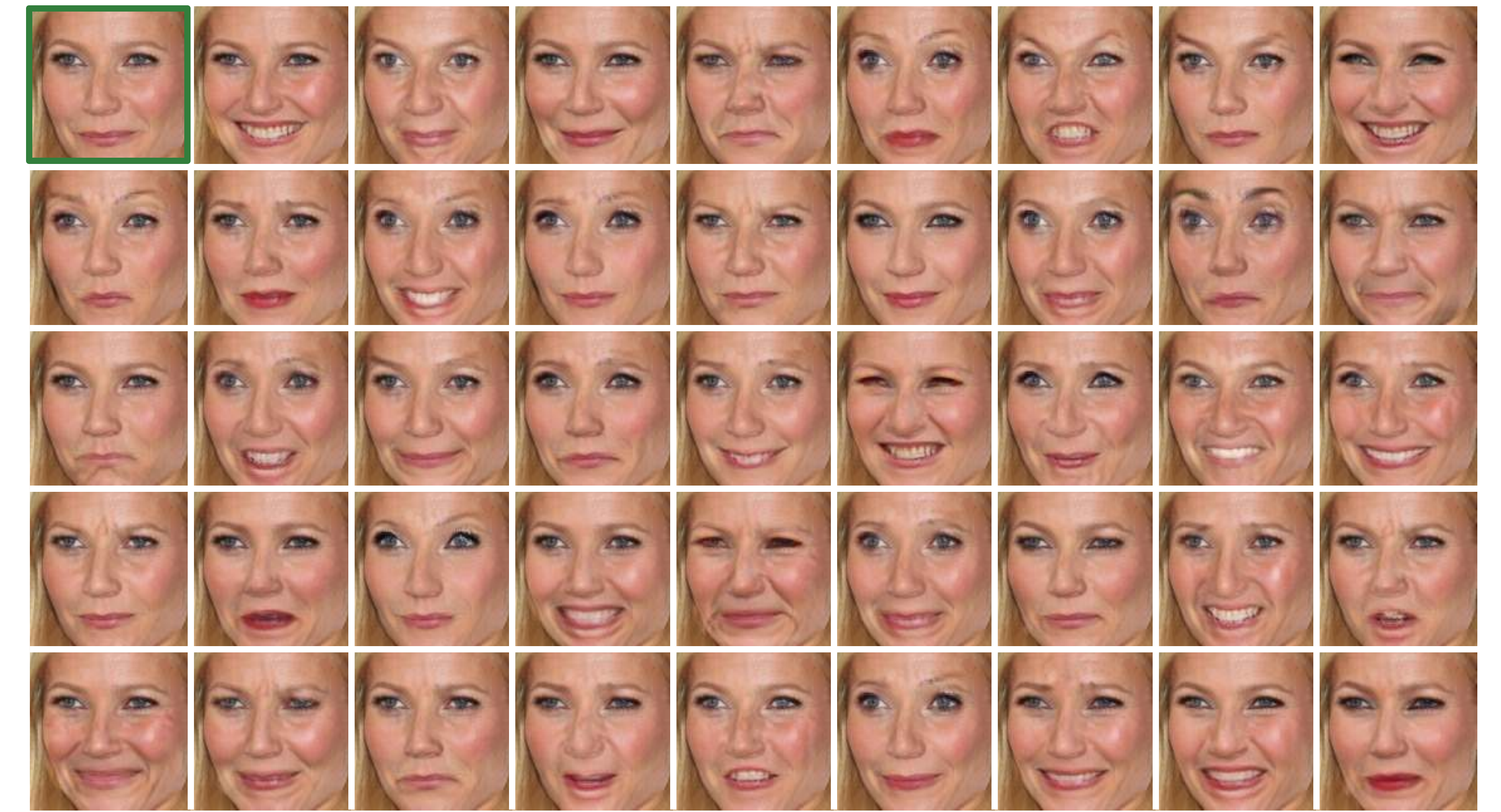}
\vspace{-7mm}
\caption{\textbf{Sampling the face expression distribution space.} As a result of applying our AU-parametrization through  the vector $\by_g$, we can synthesize, from the same source image $\bI_{\by_r}$, a large variety of photo-realistic images. }
\vspace{-4mm}
\label{fig:variability}
\end{figure}

Our model differs from these approaches in two main aspects. First, we do not condition the model on discrete emotions categories, but we learn a basis of anatomically feasible warps that allows generating a continuum of expressions.   Secondly, the use of the attention mask allows applying the transformation only on the cropped face, and put it back onto the original image without producing any artifact. As shown in Fig.~\ref{fig:state-of-the-art-comp}, besides estimating more visually compelling images than other approaches, this results on images of higher  spatial resolution.

\subsection{High Expressions Variability}
Given a single image, we next use our model to produce a wide range of   anatomically feasible face expressions while conserving the person's identity. In Fig.~\ref{fig:variability} all faces are the result of conditioning the input image in the top-left corner with a desired face configuration defined by only 14 AUs. Note the large variability of anatomically feasible expressions that can be synthesized with only 14 AUs.

\subsection{Images in the Wild}
\label{sec:wild}
\begin{figure}[t!]
\centering
\includegraphics[width=\linewidth]{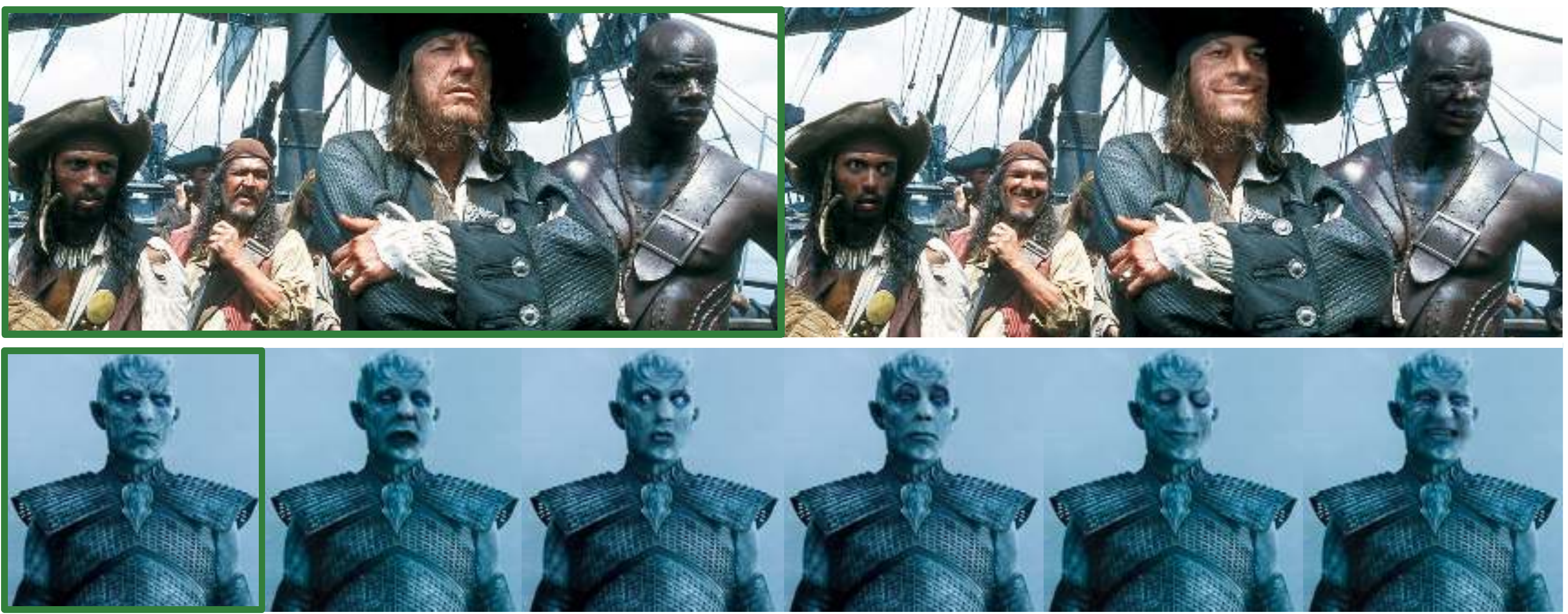}
\vspace{-7mm}
\caption{\textbf{Qualitative evaluation on images in the wild.} \textbf{Top:} We represent an image (left) from the film ``{\em Pirates of the Caribbean}'' and an its generated image obtained by our approach (right). \textbf{Bottom:} In a similar manner, we use an image frame (left) from the series ``{\em Game of Thrones}'' to synthesize   five new images   with different expressions.}
\vspace{-4mm}
\label{fig:wild}
\end{figure}

As previously seen in Fig.~\ref{fig:au_editing2}, the attention mechanism not only learns to focus on specific areas of the face but also allows merging the original and generated 
image background. This allows our approach to be easily applied to images in the wild while still obtaining   high resolution images. For these images we follow the detection and cropping scheme we described before. 
Fig.~\ref{fig:wild} shows two examples on these challenging images. Note how the attention mask allows for a smooth and unnoticeable merge between the entire frame and the generated faces.

\subsection{Pushing the Limits of the Model}
\begin{figure}[t!]
\includegraphics[width=\linewidth]{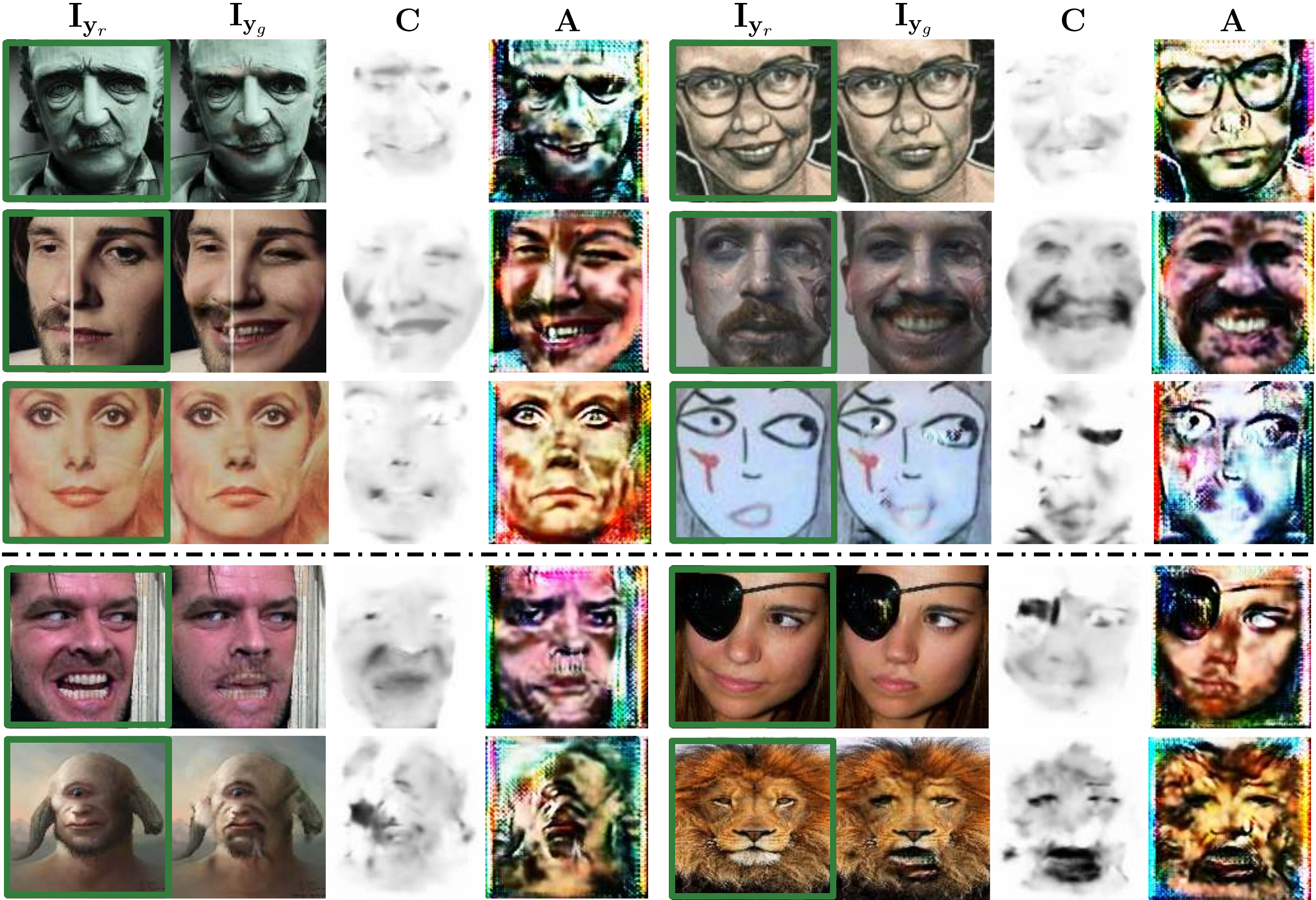}
\vspace{-7mm}
\caption{\textbf{Success and Failure Cases.} In all cases, we represent the source image $\bI_{\by_r}$, the target one $\bI_{\by_g}$, and the color and attention masks $\bC$ and $\bA$, respectively. \textbf{Top:} Some success cases in extreme situations. \textbf{Bottom:} Several failure cases.}
\label{fig:success_and_failure_cases}
\vspace{-4mm}
\end{figure}

We next push the limits of our network and discuss the model limitations.  We have   split success cases into six categories which we summarize in Fig.~\ref{fig:success_and_failure_cases}-top. The first two examples (top-row) correspond to human-like sculptures and non-realistic drawings. In both cases, the generator is able to maintain the artistic effects of the original image. Also, note how the attention mask ignores artifacts such as the pixels occluded by the glasses. The third example shows robustness to non-homogeneous textures across the face. 
Observe that the model is not trying to homogenize the texture by adding/removing the beard's hair. The middle-right category relates to anthropomorphic faces with non-real textures. As for the Avatar image, the network is able to warp the face without affecting its texture. The next category is related to non-standard illuminations/colors for which the model has already been shown robust in   Fig.~\ref{fig:intro}. The last and most surprising category is face-sketches (bottom-right). Although the generated face suffers from  some artifacts, it is still impressive how  the proposed method is still capable of finding sufficient  features on the face   to transform its expression   from worried to excited. The second case shows failures with non-previously seen occlusions such as an eye patch causing artifacts in the missing face attributes.

We have also categorized the failure cases in Fig.~\ref{fig:success_and_failure_cases}-bottom, all of them presumably due to insufficient  training data. The first case is related to errors in the attention mechanism when given extreme input expressions. The attention does not weight sufficiently the color transformation causing transparencies. 

 The model also fails when dealing with non-human anthropomorphic distributions as in the case of cyclopes. Lastly, we tested the model behavior when dealing with animals and observed   artifacts like human   face features.

\section{Conclusions}
We have presented a novel GAN model for   face animation in the wild that can be trained in a fully unsupervised manner. It advances   current works which, so far, had only addressed the problem for discrete emotions category editing and portrait images. Our model encodes anatomically consistent face deformations parameterized by means of AUs. Conditioning the GAN model on  these AUs allows the generator to render a wide range of expressions by simple interpolation. Additionally, we embed an attention model within the network which allows focusing  only on those regions of the image relevant for every specific expression. By doing this, we can easily process images in the wild,   with distracting backgrounds and illumination artifacts. We have exhaustively evaluated the model capabilities and limits in the EmotioNet~\cite{benitez2016emotionet} and RaFD~\cite{langner2010presentation} datasets as well as in images   from movies. The results are very promising, and show smooth transitions between different expressions. This opens the possibility of applying our approach to video sequences, which we plan to do in the future. 

\vspace{1mm}
\noindent\textbf{Acknowledgments:} This work is partially supported by the Spanish Ministry of Economy and Competitiveness under projects HuMoUR TIN2017-90086-R, ColRobTransp DPI2016-78957 and Mar\'ia de Maeztu Seal of Excellence MDM-2016-0656; by the EU project AEROARMS ICT-2014-1-644271; and by the Grant R01-DC- 014498 of the National Institute of Health. We also thank Nvidia for hardware donation under the GPU Grant Program.

\bibliographystyle{splncs04}
\bibliography{egbib}
\end{document}